\documentclass[letterpaper, 10 pt, conference]{ieeeconf}  

\IEEEoverridecommandlockouts                              

\onecolumn

\usepackage{graphics} 
\usepackage{graphicx}

\usepackage{epsfig} 
\usepackage{amsmath} 
\usepackage{amssymb}  
 \usepackage{algorithm}
\usepackage{algpseudocode}
\algnewcommand\AAND{\textbf{ and }}
\algnewcommand\Or{\textbf{ or }}
\usepackage{color}
\usepackage{citesort}
\usepackage{flushend}
\usepackage{url}

\DeclareMathAlphabet{\pazocal}{OMS}{zplm}{m}{n}

\usepackage{array}
\newcolumntype{C}[1]{>{\centering\arraybackslash}p{#1}}
\newcolumntype{M}[1]{>{\raggedright\arraybackslash}p{#1}}

\usepackage{array} 
\newcolumntype{L}[1]{>{\raggedright\let\newline\\\arraybackslash\hspace{0pt}}m{#1}}	
\newcolumntype{S}[1]{>{\centering\let\newline\\\arraybackslash\hspace{0pt}}m{#1}}
\newcolumntype{R}[1]{>{\raggedleft\let\newline\\\arraybackslash\hspace{0pt}}m{#1}}

\algnewcommand\pushup{\vspace{-1ex}}
\algnewcommand\pushuphalf{\vspace{-0.5ex}}

\makeatletter
\renewcommand*{\@opargbegintheorem}[3]{\trivlist
  \item[\hskip \labelsep{\itshape #1\ #2}] \textit{(#3)}\ }
\makeatother

\title{\LARGE \bf
Visual-Thermal Camera Dataset Release and Multi-Modal Alignment without Calibration Information
}

\author{Frank Mascarich$^{1}$, Kostas Alexis$^{2}$
\thanks{The authors are with the Autonomous Robots Lab, $^{1}$University of Nevada, Reno \& $^{2}$NTNU - Norwegian University of Science and Technology        {\tt\small fmascarich@nevada.unr.edu, konstantinos.alexis@ntnu.no}}%
}

\begin{document}

\maketitle
\thispagestyle{empty}
\pagestyle{empty}

\begin{abstract}
This report accompanies a dataset release on visual and thermal camera data and details a procedure followed to align such multi-modal camera frames in order to provide pixel-level correspondence between the two without using intrinsic or extrinsic calibration information. To achieve this goal we benefit from progress in the domain of multi-modal image alignment and specifically employ the Mattes Mutual Information Metric to guide the registration process. In the released dataset we release both the raw visual and thermal camera data, as well as the aligned frames, alongside calibration parameters with the goal to better facilitate the investigation on common local/global features across such multi-modal image streams. 
\end{abstract}

\section{INTRODUCTION}

Autonomous robotic operation and scene understanding often requires the fusion of diverse sensing modalities. Especially when robot navigation and environment cognition in visually-degraded environments is considered~\cite{ARTIFACTS_AEROCONF2020,GBPLANNER_IROS_2019,sizintsev2019multi}, robotic systems may benefit if equipped with not only traditional camera systems but also sensors with complementary capabilities. Thermal vision in particular offers certain capabilities of interest including the fact that it can remain informative in conditions of darkness or certain types of obscurants such as dust. In previous work, the research community has investigated - among others - the benefits of thermal vision specifically for the problem of thermal-inertial odometry estimation~\cite{KTIO_ICRA_2019,VTIMUDVE_AEROCONF_2019,khattak2019robust,zhao2020tp,khattak2018marker} and object detection~\cite{treptow2006real,ciric2013computationally,yin2006moving}. When multiple modalities are available onboard an essential question relates to the best way that their data may be fused in order to allow maximum resilience and information-rich robotic operation. Motivated by the above, we seek to identify how to best fuse the data from visual and thermal vision. Towards that goal and with the aim to investigate both hand-designed and data-driven approaches for the underlying multi-modal data association problem, there is need to develop datasets that allow to study the cross-modality relations of visual and thermal cameras for robotic applications. 

This working paper serves to accompany a dataset release on visual and Long-Wave InfraRed (LWIR) camera data and to further demonstrate a process for their alignment assuming lack of intrinsic and extrinsic calibration parameters, alongside no knowledge of the depth associated with each pixel. The method employed considers the fact that visual and thermal camera are different in nature and thus the image intensities do not necessarily present high similarity - especially in a local sense. Reflecting this fact, the employed approach to achieve alignment is a multi-modal one and in particular it relates to the utilization of the Mattes Mutual Information Criterion in order to guide the alignment. Given the two image sets, visual and thermal, the method seeks to identify the geometric transformation $\mathbf{g}$ between each pair that maps every point $\mathbf{x}$ in the visual frame to the point $\mathbf{g}(\mathbf{x})$ on the thermal image. Notably, the dataset release does contain calibration information. 

\section{MULTI-MODAL IMAGE ALIGNMENT}\label{sec:perception}

\begin{figure}[h]
\centering
    \includegraphics[width=0.75\columnwidth]{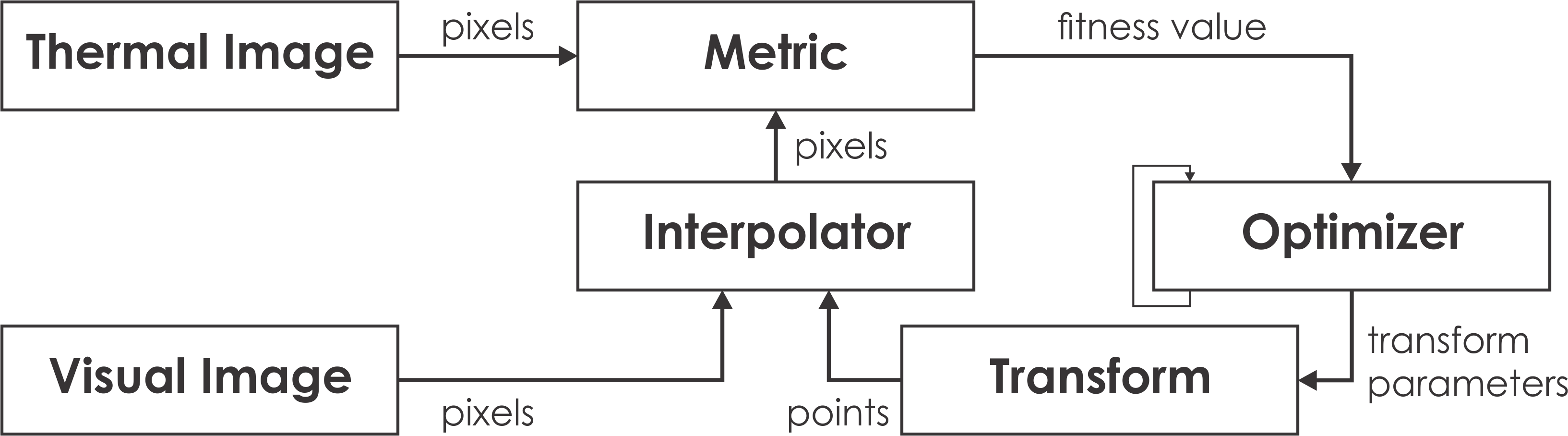}
\caption{Block diagram overview of the multi-modal image registration process utilized. }\label{fig:multimodalalignment}
\end{figure}

In this section, the basic steps and decision choices for the considered visual-thermal alignment process are detailed. This involves the selection of the transformation type, interpolation scheme and most importantly the employed mutual information criterion and the optimization process involved. The outline of the registration process is depicted in Figure~\ref{fig:multimodalalignment}.

%
%

\subsection{Transformation}

To allow satisfactory image alignment between visual and thermal frame pairs without access to calibration information we allowed to test different transformation options. In particular, we evaluated the optimization process when utilizing the degrees of freedom of two transformation options, namely similarity transformation type and affine transformation type. We found similar results with either of the two. The affine transformation type allows for translation, rotation, scaling and shearing. The similarity transformation type allows for translation, rotation and scale. Both of them always involve nonreflective transformations. Below we outline the individual operations of the affine transformation:

\begin{eqnarray}
\begin{bmatrix}
1 & 0 & 0\\ 
0 & 1 & 0\\ 
t_x & t_y & 1
\end{bmatrix}, ~
\begin{bmatrix}
s_x & 0 & 0\\ 
0 & s_y & 0\\ 
0 & 0 & 1
\end{bmatrix}, ~
\begin{bmatrix}
1 & sh_y & 0\\ 
sh_x & 1 & 0\\ 
0 & 0 & 1
\end{bmatrix}, ~
\begin{bmatrix}
\cos q & \sin q & 0\\ 
-\sin q & \cos q & 0\\ 
0 & 0 & 1
\end{bmatrix}
\end{eqnarray}
where $t_x,t_y$ specify the displacement in the $x,y$ axes respectively, $s_x,s_y$ are the scale factors along those axes, $sh_x,sh_y$ are the shear factors along $x,y$, and $q$ specifies the angle of rotation. The results presented in Section~\ref{sec:exp} are all using similarity transformation.

\subsection{Interpolation}

The transformations can lead to points in coordinates on non-grid positions of the image (e.g., non-integer pixel coordinates). In order to allow the comparison of the transformed image with the other image modality, intensity values can be calculated via interpolation. A variety of possible options are available to act as means of interpolation including nearest-neighbor and linear interpolation schemes. In this work, spline interpolation is utilized~\cite{schoenberg1988contributions}. Splines are piecewise-applied low-order polynomials that present a smooth transition. Using low-order polynomials ensures that the computational requirements remain relatively low. At the same time, undesired oscillations resulting from the possible use of high-order polynomials are also avoided. 

\subsection{Multi-Modal Image Alignment}

To align the considered multi-modal data from visual and thermal images we seek to identify the transformation parameters $\boldsymbol{\mu}$ that maximize an image similarity function $S$:

\begin{eqnarray}
\boldsymbol{\mu}_{opt} = \underset{\boldsymbol{\mu}}{\mathrm{argmax}}~S(\boldsymbol{\mu})
\end{eqnarray}
where for the purposes of this work, mutual information is used as the image similarity function. The problem as stated is a maximization problem but in turn we minimize the negative of the function $S$, between the visual image and the transformed thermal image, which considering the Mattes Mutual Information Criterion takes the form: 

\begin{eqnarray}
S({\boldsymbol{\mu}}) = -\sum_{\iota \in L_T} \sum_{\kappa \in L_V} p(\iota, \kappa; \boldsymbol{\mu})\log \frac{p(\iota,\kappa;\boldsymbol{\mu})}{p_T(\iota;\boldsymbol{\mu})p_V(\kappa)}
\end{eqnarray}
where $p,p_T,p_V$ are the joint, marginal thermal camera data, and marginal visual camera data probability distributions, while $L_T$ and $L_V$ are the intensity values of the test and reference images.

As described in~\cite{mattes2001nonrigid} the probability distributions used to compute mutual information are based on marginal and joint histograms of the two types of image modalities. In order to form continuous estimates of the underlying image histograms, Parzen windowing can be utilized. In such a manner, the joint distribution becomes an explicitly differentiable function. Let $\beta^{(3)}$ be a cubic spline Parzen window and similarly $\beta^{(0)}$ be a zero-order spline Parzen window then the joint probability takes the form: 

\begin{eqnarray}
p(\iota, \kappa;\boldsymbol{\mu}) = a \sum_{\mathbf{x}\in V}\beta^{(0)}\left ( \kappa - \frac{f_V(\mathbf{x})-f_V^\prime}{\Delta b_V} \beta^{(3)} \left ( \iota - \frac{f_T (g(\mathbf{x};\boldsymbol{\mu}))-f_T^\prime}{\Delta b_T} \right ) \right )
\end{eqnarray}
where $a$ is a normalization factor ensuring that $\sum p(\iota, \kappa)=1$ and $f_V(\mathbf{x}),f_T(g(\mathbf{x};\boldsymbol{\mu}))$ are samples of the two modality images (interpolated) respectively with $\mathbf{x}=[x,y,z]^T$ being any voxel location in the visual image (coordinates and intensity), while $\Delta b_T, \Delta b_V$ are the intensity range of each bin (used to to fit into a predefined number of bins in the intensity distribution). Each contribution is normalized by the minimum intensity value, denoted as $f_V^\prime, f_T^\prime$, as well as the width of the histogram bin. The summation range $V$ is the set of voxel pairs that contribute to the distribution. 

Subsequently, the marginal discrete probability for each of the thermal camera images is computed from the joint distribution:

\begin{eqnarray}
p_T(\iota;\boldsymbol{\mu}) = \sum_{\kappa \in L_V}p(\iota, \kappa;\boldsymbol{\mu})
\end{eqnarray}

Respectively, the visual camera data marginal distribution takes the form: 

\begin{eqnarray}
p_V(\kappa) = a\sum_{\mathbf{x}\in V}\beta^{(0)}\left ( \kappa - \frac{f_V(\mathbf{x})-f_V^\prime}{\Delta b_V} \right )
\end{eqnarray}
For further details the interested reader is referred to published work and relevant software tools~\cite{mattes2001nonrigid,mccormick2014itk,pluim2003mutual,klein2007evaluation}.

In order to robustly solve this registration problem we employ the $(1+1)$ Evolutionary Algorithm (EA). The $(1+1)$ EA method is a nonlinear optimization algorithm which does not utilize gradient information but is based on a probabilistic, evolutionary strategy~\cite{styner2000parametric,schwefel1993evolution}. The algorithm operates on one parent and one descendant at a time. From the current transformation (parent), a new transformation (descendant) is generated via small modifications (mutation). If it turns out that the new transformation yields a higher similarity (fitness) score then it gets selected as a the new current transformation. In this process, the generation of a new solution is controlled by a gaussian probability function the center of which is at the current location in the search-space. To best guide the evolutionary process, this probability function shrinks if the new transformation yields a lower similarity score and grows if the new transformation yields a higher similarity score than the current one. A set of parameters allow to tune the behavior of the $(1+1)$ algorithm including the growth factor, the shrink factor, the initial radius and the epsilon value.

\subsection{Multi-Resolution Image Pyramids}

In principle, the aforementioned registration process may benefit by utilizing optimization across the scales of an image pyramid. However, for the purposes of this work - and at least the used data - it was found that employing pyramid levels corresponding to reduced resolutions was not beneficial. This may be attributed to the fact that thermal camera data lack in terms of sharpness as compared to visual data. This was especially observed in those datasets involving surfaces at close proximity and with more flat thermal gradients. 

\section{VISUAL THERMAL DATASETS}\label{sec:planning}

Two sets of datasets are released, namely a) automotive data integrating visual and thermal cameras and operating both in daytime and nighttime conditions, as well as b) subterranean aerial robotic scouts operating in underground mines and tunnels. A brief overview of the datasets is presented below. 

\noindent \textbf{Car-Daytime:} In this dataset, a Lincoln MKZ vehicle owned by the Nevada Center for Applied Research (NCAR) was utilized and retrofitted with visual and thermal cameras. The sensor specifics relating to this study are summarized in Table~\ref{tab:datasets}, while the environment of the dataset relates to areas of the city of Reno and specifically around the University. 

\noindent \textbf{Car-Nighttime:} In this dataset the same vehicle as above was utilized and thus the same sensor parameters apply. The environment is similar with the most notable change relating to the fact that in this case nighttime operation is considered. 

\noindent \textbf{Aerial Scout-Tunnel:} In this dataset, an aerial robotic scout was utilized and operated to explore an abandoned train tunnel. The system integrates visual and thermal cameras as detailed in Table~\ref{tab:datasets}. 

\noindent \textbf{Aerial Scout-Mine:} In this dataset, an aerial robotic scout is used to explore an underground mine. The system integrates the sensors detailed in Table~\ref{tab:datasets}. 

\begin{table}[h]
\begin{center}

\begin{tabular}{|l|l|l|l|l|}
\hline
\textbf{Dataset}       & \textbf{Visual Camera}        & \textbf{Lens} & \textbf{Thermal Camera} & \textbf{Lens}                                                        \\ \hline
Car-Daytime            & Point Grey C3-U3-13Y3M        & $3.6\textrm{mm}, \textrm{F}2.0$ & FLIR Boson              & $95^\circ \textrm{HFOV}, 4.9\textrm{mm}$                                               \\ \hline
Car-Nighttime          & Point Grey C3-U3-13Y3M        & $3.6\textrm{mm}, \textrm{F}2.0$ & FLIR Boson              & $95^\circ \textrm{HFOV}, \textrm{4.9mm}$                                               \\ \hline
Aerial Scout-Tunnel & FLIR Blackfly BFS-U3-16S2C-CS & $3.6\textrm{mm}, \textrm{F}2.0$ & FLIR Tau2               & $90^\circ \textrm{HFOV}, 7.5\textrm{mm}$                                               \\ \hline
Aerial Scout-TRJV    & Point Grey C3-U3-13Y3M        & $3.6\textrm{mm}, \textrm{F}2.0$ & FLIR Tau2               & $90^\circ \textrm{HFOV}, 7.5\textrm{mm}$ \\ \hline
\end{tabular}
\caption{Sensors used in the considered experiments.}\label{tab:datasets}
\end{center}
\end{table}

%
\begin{figure}[h]
\centering
    \includegraphics[width=0.99\columnwidth]{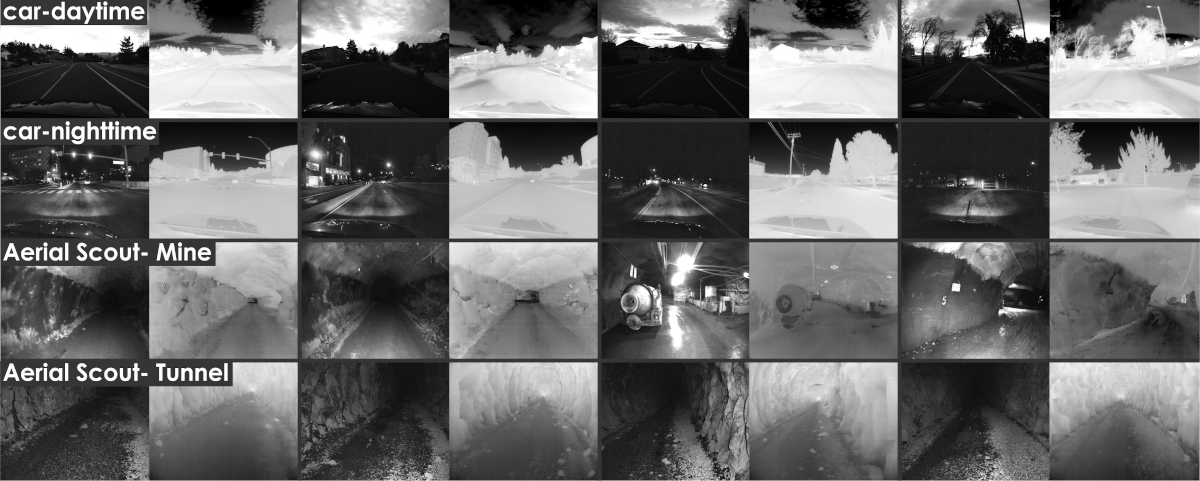}
\caption{Indicative images from the raw visual and thermal camera frames of the released dataset. Top row: Car-Daytime dataset, second row: Car-Nighttime dataset, third row: Aerial Scout-Mine dataset, bottom row: Aerial Scout-Tunnel dataset.}\label{fig:datasetraw}
\end{figure}
%

\begin{figure}[h]
\centering
    \includegraphics[width=0.99\columnwidth]{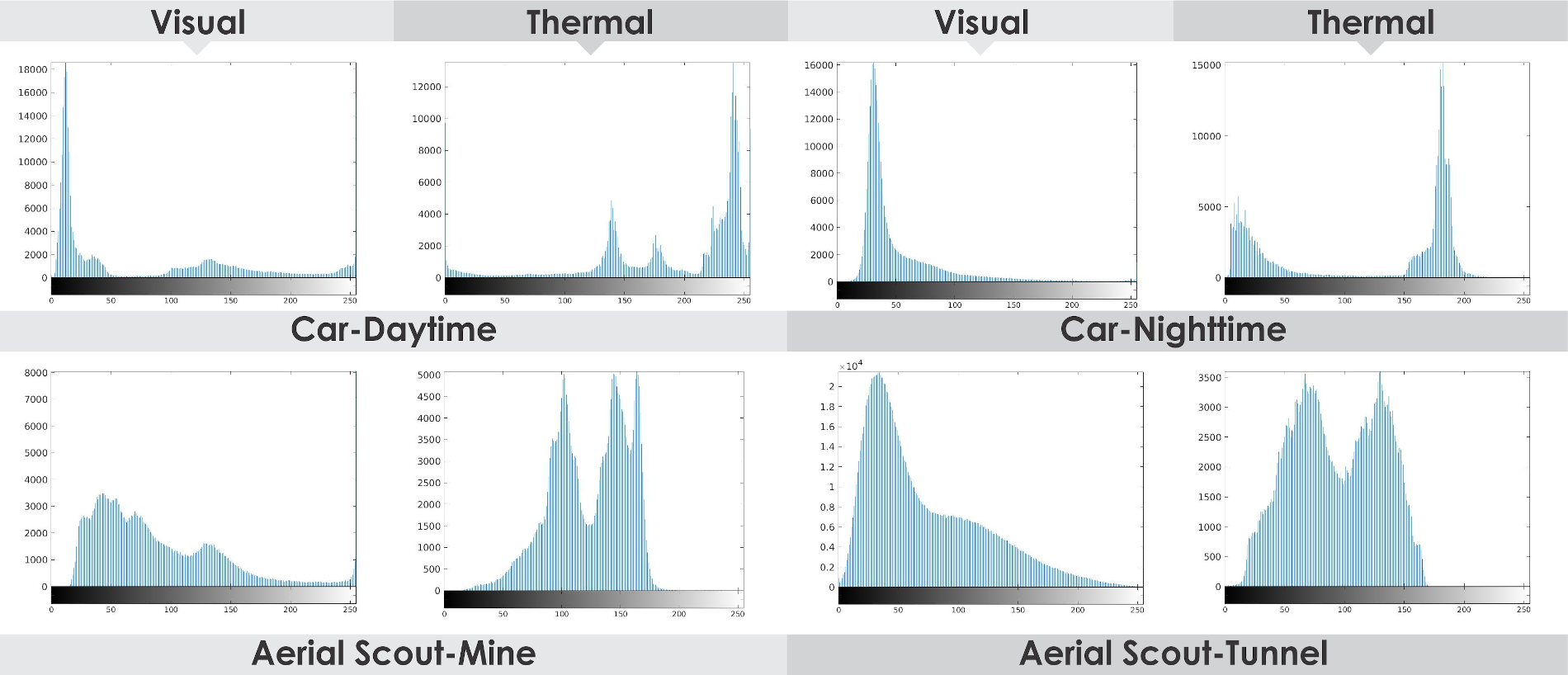}
\caption{Indicative histogram plots ($256$ bins) for visual and thermal camera data from the released datasets.}\label{fig:histogramsraw}
\end{figure}

To allow to better understand the information content of the different camera modalities, especially as this relates to the utilized mutual information metric, Figure~\ref{fig:histogramsraw} presents indicative histograms for visual and thermal camera frames from the released datasets.

%
%

Overall, the released set of datasets (available at \url{https://www.autonomousrobotslab.com/vtdataset.html}) contains for each case the following data: a) raw visual camera frames, b) associated raw thermal camera frames, c) the histograms for both types of frames, c) the aligned multi-modal frames using the methodology outlined, and d) calibration information for the cameras utilized. It is noted that not all the recorded frames are released but a rather low framerate is provided for all datasets. This is due to the fact that the purpose of this dataset is not about odometry estimation. Other such datasets have been previously released (e.g., \url{https://www.autonomousrobotslab.com/ktio.html}).

\section{EXPERIMENTAL VERIFICATION}\label{sec:exp}

In this section, indicative frames of the datasets are presented alongside results of the visual-thermal alignment process. The complete datasets including the raw visual and thermal camera frames, as well as the aligned result for every visual-thermal camera pair are released and can be accessed at~\url{https://www.autonomousrobotslab.com/vtdataset.html}.

\begin{figure}[h!]
\centering
    \includegraphics[width=0.99\columnwidth]{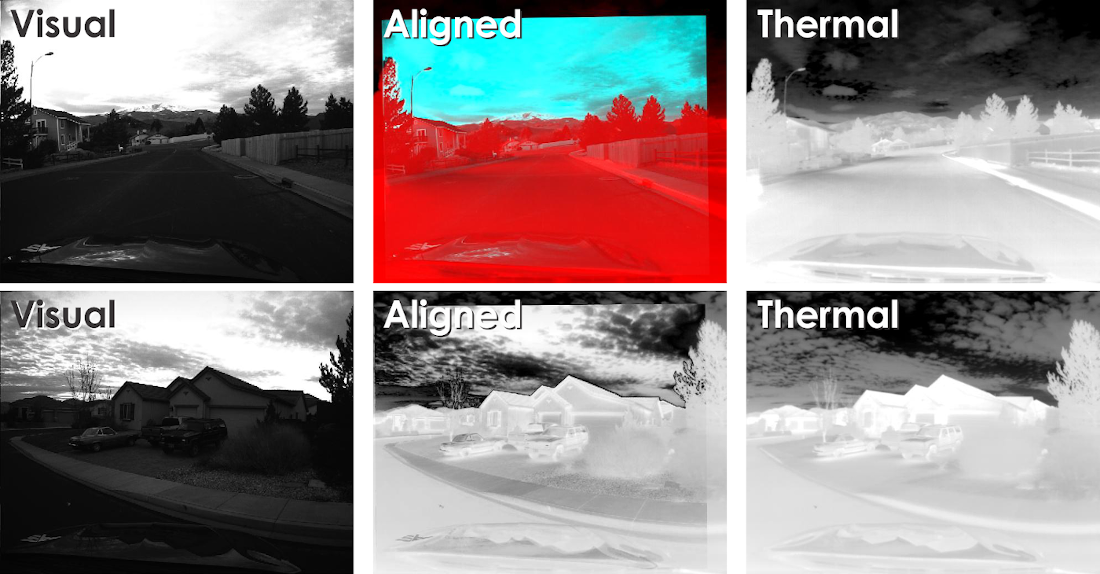}
\caption{Indicative results of the considered visual-thermal camera frames alignment process assuming no access to calibration or depth information for the case of the Car-Daytime dataset. }\label{fig:car}
\end{figure}

Figure~\ref{fig:car} presents two tuples of visual and thermal camera data, alongside the result of their alignment for the ``Car-Daytime'' dataset. Two different colorization methods are used for the aligned result, namely using ``red-cyan'' color channels and monochromatic ``difference''. Careful observation of the images allows to understand that the image alignment is generally correct with specific features such as the roadside light bulb, or the corners of the building roofs appearing aligned for the visual and thermal camera data. 

%
%

\begin{figure}[h!]
\centering
    \includegraphics[width=0.99\columnwidth]{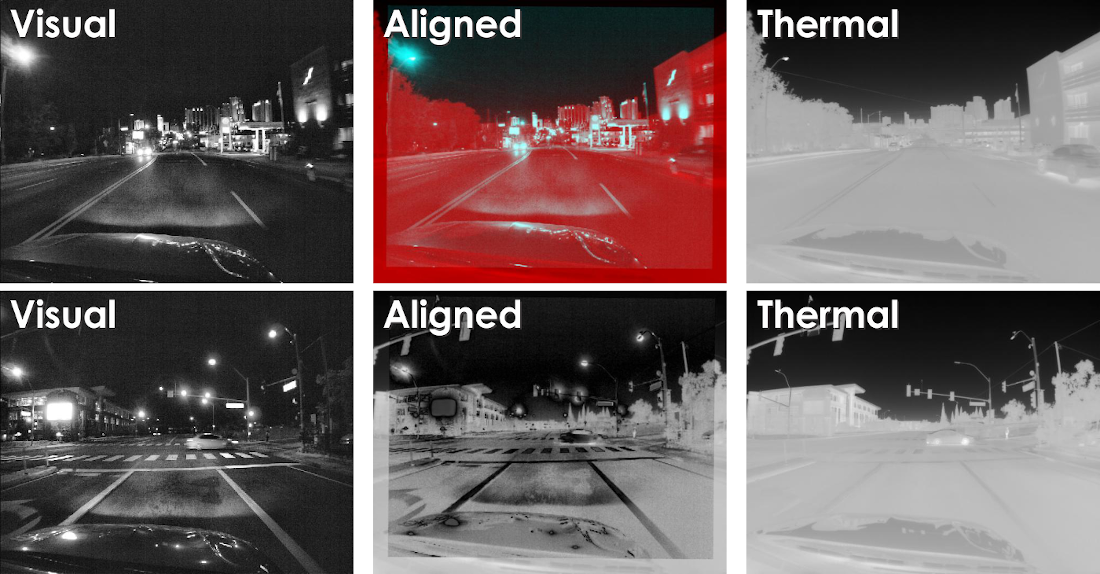}
\caption{Indicative results of the considered visual-thermal camera frames alignment process assuming no access to calibration or depth information for the case of the Car-Nighttime dataset. }\label{fig:car_night}
\end{figure}

Figure~\ref{fig:car_night} presents two tuples of visual and thermal camera data, alongside the result of their alignment for the ``Car-Nighttime'' dataset. Careful observation of the images allows to understand that the image alignment is generally correct with specific features such as the roadside light bulb, the buildings and the large screen appearing aligned for the visual and thermal camera data. 

%
%

\begin{figure}[h!]
\centering
    \includegraphics[width=0.99\columnwidth]{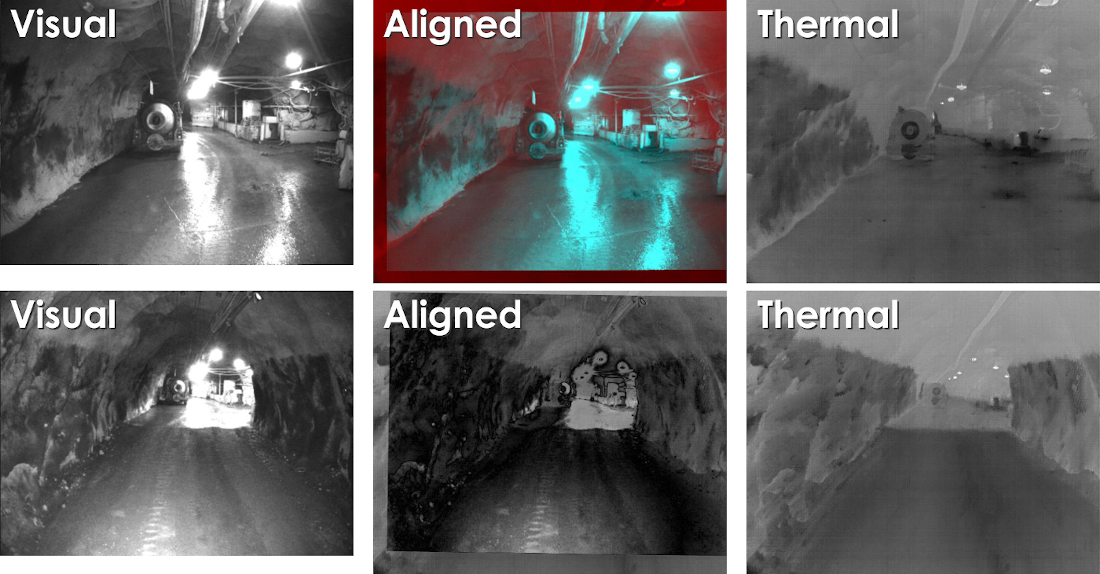}
\caption{Indicative results of the considered visual-thermal camera frames alignment process assuming no access to calibration or depth information for the case of the Aerial Scout-Mine dataset. }\label{fig:trjv}
\end{figure}

Figure~\ref{fig:trjv} presents two tuples of visual and thermal camera data, alongside the result of their alignment for the ``Aerial Scout-Mine'' dataset. Careful observation of the images allows to understand that the image alignment is generally correct with specific features such as the light bulbs or the machinery on the left of the images being aligned for the visual and thermal camera data (but with reduced quality compared to the previous dataset). 

%
%

\begin{figure}[h!]
\centering
    \includegraphics[width=0.99\columnwidth]{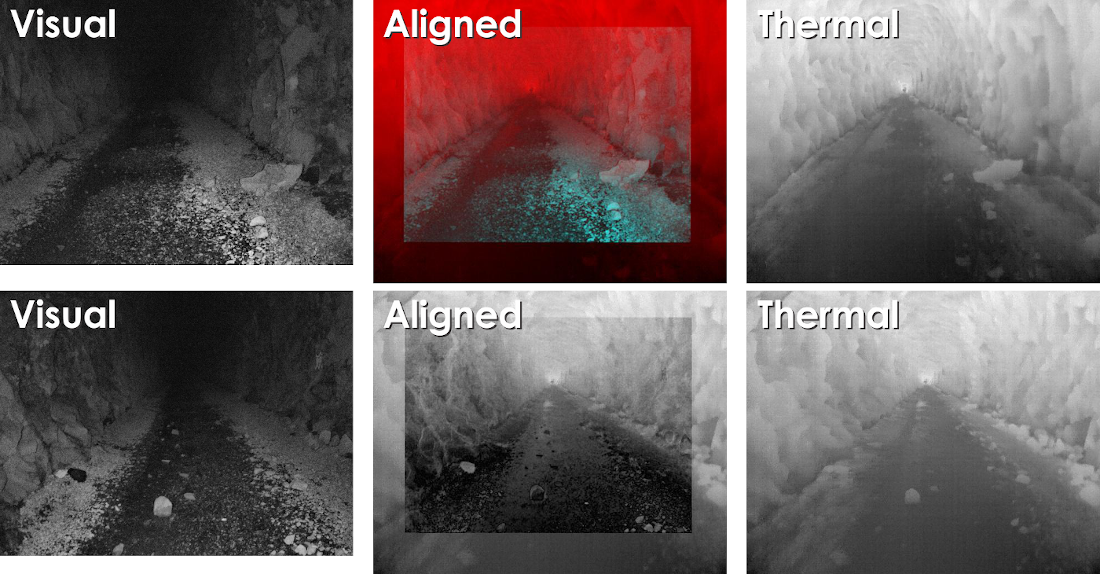}
\caption{Indicative results of the considered visual-thermal camera frames alignment process assuming no access to calibration or depth information for the case of the Aerial Scout-Tunnel dataset. }\label{fig:truckee}
\end{figure}

Figure~\ref{fig:truckee} presents two tuples of visual and thermal camera data, alongside the result of their alignment for the ``Aerial Scout-Tunnel'' dataset. Careful observation of the images allows to understand that the image alignment is generally correct with specific features such as the rocks on the ground appearing aligned for the visual and thermal camera data (but with reduced quality compared to the firs two dataset cases). 

%
%

\begin{figure}[h!]
\centering
    \includegraphics[width=0.99\columnwidth]{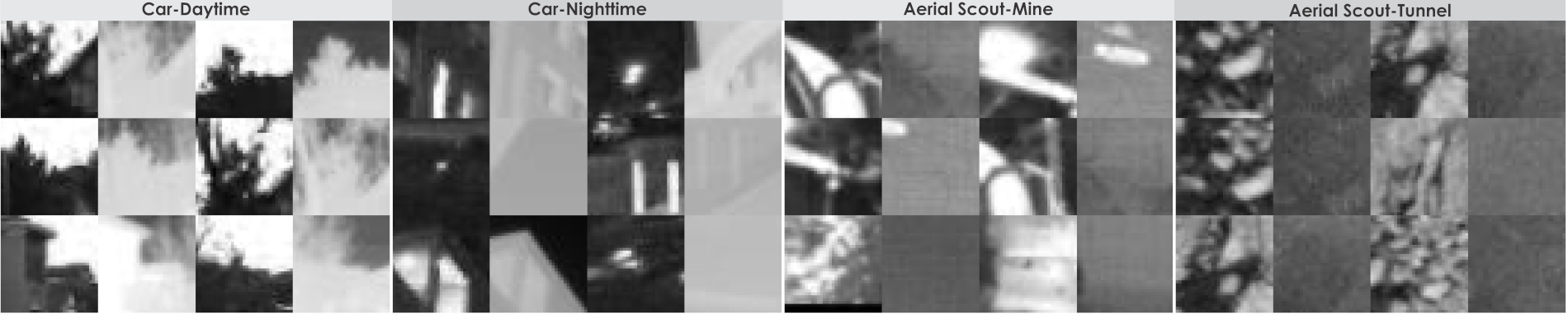}
\caption{Indicative $32\times 32$ image patches from visual and thermal data with the centroid of each visual patch being the coordinates of a FAST detection and the centroid of the thermal patch acquired through the identified alignment transformation. One image pair for each of the datasets is considered for this result and only six (randomly selected) detected corners are presented for each image pair. Naturally, datasets that are rich both visually and thermally allow for rather more ``matching'' multi-modal frames. }\label{fig:localfeatures}
\end{figure}

The derived results correspond to the demonstration of a possible approach for visual-thermal camera data alignment assuming no access to calibration files and depth information. It is highlighted that the released dataset does contain calibration information. However, it was considered that given the lack of large-scale datasets for combined visual and thermal camera data, then the need to utilize as much of the data available as possible would mean that outlining a process for their uncalibrated alignment is of practical utility. The quality of the registration is not always excellent but largely allows for basic alignment results at least via a qualitative assessment. Figure~\ref{fig:localfeatures} presents indicative results from the process of performing FAST corner extraction on an indicative visual frame from each of the datasets and identifying the respective locations on the associated thermal image. A random small subset of the detected corners and $32\times 32$ patches around them are presented (six per image pair). When objects are very close to the camera frame then naturally the methodology suffers as at that point, access to calibration information is even more essential. Other cases that introduce worse alignment results also exist and are evident in particular in the underground datasets where both visual and thermal information content is less rich. However, it is important to mention that out of evaluating different approaches including traditional feature matching (e.g., using SURF), the methodology to employ the mutual information criterion seemed to provide the most robust and reliable alignment. Finally, an additional step could be employed that relates to the identification of a non-rigid transformation to best align the visual and thermal frames pixel-wise. This will naturally deform the images and will introduce a non-rigid change that is unrealistic. It may however allow for locally improved alignments and thus can be considered as a step to be applied in a local manner for better local exploitation of the dataset. 

%
%

\section{CONCLUSIONS}\label{sec:conclusions}

This working report serves to accompany a dataset release on visual and thermal camera data and further present an algorithmic process for the multi-modal alignment of these data assuming no access to intrinsic or extrinsic calibration parameters and no knowledge of the depth information for any of the image pixels. The strategy utilized for this multi-modal alignment process involves the use a mutual information metric to guide the registration process and achieve good pixel-to-pixel matching. The released dataset involves both urban and underground scenes in diverse light conditions and incorporates calibration information. 

\bibliographystyle{IEEEtran}
\bibliography{VT_Align}

\end{document}